\documentclass[preprint,12pt,3p]{elsarticle}

\usepackage{blindtext, graphicx, float, placeins, dblfloatfix, url, caption}
\usepackage{caption}
\usepackage{subcaption}
\usepackage{algorithm,algorithmic}
\usepackage{amssymb}

\begin{document}

\begin{frontmatter}

\title{Weight Initialization of Deep Neural Networks(DNNs) using Data Statistics}

\author{Saiprasad Koturwar}
\address{Department of Electrical Engineering\\
          Indian Institute of Technology Bombay\\
          Powai, Mumbai -- 400076\\}

\author{Shabbir N Merchant}
\address{Department of Electrical Engineering\\
          Indian Institute of Technology Bombay\\
          Powai, Mumbai -- 400076\\}
\ead{merchant@ee.iitb.ac.in}
\ead{s.koturwar@gmail.com}

\begin{abstract}
Deep neural networks (DNNs) form the backbone of almost every state-of-the-art technique in the fields such as computer vision, speech processing and text analysis. The recent advances in computational technology have made the use of DNNs more practical. Despite the overwhelming performances by DNN and the advances in computational technology, it is seen that very few researchers try to train their models from the scratch. Training of DNNs still remains a difficult and tedious job. The main challenges that researchers face during training of DNNs are the vanishing/exploding gradient problem and the highly non-convex nature of the objective function which has upto million variables. The approaches suggested in He and Xavier solve the vanishing gradient problem by providing a sophisticated initialization technique. These approaches have been quite effective and have achieved good results on standard datasets, but these same approaches do not work very well on more practical datasets. We think the reason for this is not making use of data statistics for initializing the network weights. Optimizing such a high dimensional loss function requires careful initialization of network weights. In this work we propose a data dependent initialization and analyse it's performance against the standard initialization techniques such as He and Xavier. We performed our experiments on some practical datasets and the results show our algorithm's superior classification accuracy. 
\end{abstract}

\begin{keyword}
Deep neural networks \sep Data dependent initialization \sep Non-convex optimization \sep DNN visualization \sep vanishing/exploding gradient problem
\end{keyword}

\end{frontmatter}


\section{Introduction}
\label{intro}
\indent Deep neural networks(DNN) are being used very heavily in the fields such as computer vision\cite{VGG}, speech processing\cite{speech} and text processing\cite{sentiment}. Each and every state-of-the-art technique in these fields make use of deep learning in one way or the other. Although deep neural networks achieve near human performances on the complex task such as object recognition\cite{VGG}, image segmentation\cite{segemnent} and localization\cite{YOLO}, most of the researchers chose not to build their network from scratch. The reason is training a DNN still remains a very challenging task. Training a DNN requires optimizing a cost function which is a highly non-convex function of the network weights. And the number of weights increase exponentially with the size of the network, even with a 3-4 layer DNN we will have almost a million variables. In such cases finding a good initialization becomes extremely important to achieve good results.
\newline \indent The data statistics provides us an useful insight into the data. The previously mentioned techniques(He and Xavier) are independent of data statistics and depend only on the size of the network. 
\newline \indent In this work, we propose an initialization technique that makes use of data statistics for weight initialization and also incorporates the conditions(i.e all layers learn at the same rate) from He and Xavier. 
\section{Related Work}
\indent \indent The problem of vanishing/exploding gradient was first identified by Sepp Hochreiter in his diploma thesis\cite{sepp}. Since then there have been few approaches to counter this problem such as unsupervised pre-training\cite{pretraining}. Although this method has achieved good results but is very prone to overfitting and involves a very lengthy process of training. The work done by Xavier\cite{glorot} and He\cite{He} solve the vanishing/exploding gradient problem by providing a sophisticated initialization technique which makes sure that all the layers of network learn at almost the same rate. These approaches perform pretty well on balanced datasets and are less complicated than unsupervised pre-training approach. But these same approaches do not work as expected on practical datasets. The following graphs depict the performance of these initialization techniques on practical and unbalanced dataset (Here we have used kaggles Statefarm distracted driver detection dataset).
\begin{figure}[h]
\centering
\begin{subfigure}{.5\textwidth}
  \centering
  \includegraphics[width=0.8\textwidth,height=0.2\textheight]{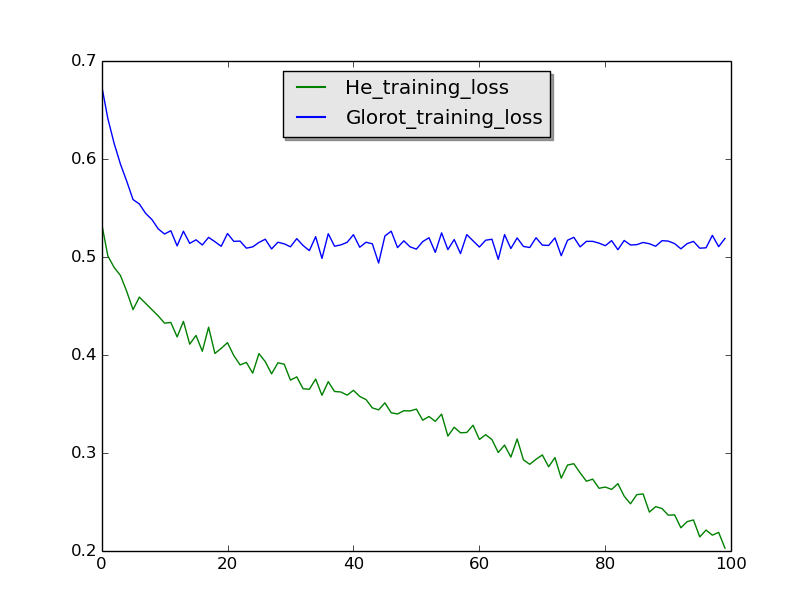}
  \caption{Training Loss}
  \label{fig:sub1}
\end{subfigure}%
\begin{subfigure}{.5\textwidth}
  \centering
  \includegraphics[width=0.8\textwidth,height=0.2\textheight]{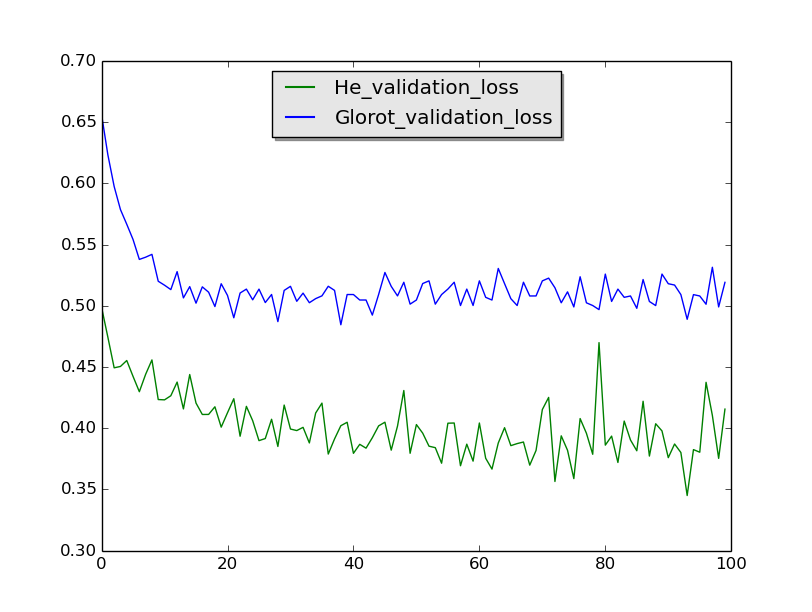}
  \caption{Validation Loss}
  \label{fig:sub2}
\end{subfigure}
\caption{Logloss vs epoch for He and Glorot initialization}
\label{fig:test}
\end{figure}
\newpage \indent As can be concluded from the above plots, these approaches do not work well on the practical datasets. Although He initialization achieves good performance on training datasets but the validation performance is very bad.
\section{Data dependent initialization}
\indent Deep neural networks learn the features through hierarchy, learning simpler features in the bottom layers and more complex features in the higher layers. This property of DNN's can be analysed through visualization of state-of-the-art network models such as VGG. The following figure shows the visualized filter at different layers. 
\begin{figure}[h]
\begin{center}    
	\begin{subfigure}[h]{6.75 cm}
		\includegraphics[width=1.5 cm,height=1.5cm]{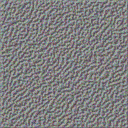}
		\includegraphics[width=1.5 cm,height=1.5cm]{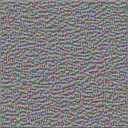}
		\includegraphics[width=1.5 cm,height=1.5cm]{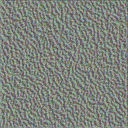}
		\includegraphics[width=1.5 cm,height=1.5cm]{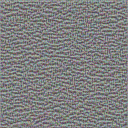}
	  \caption{Visualization of layer one filters}	
	\end{subfigure}
	\begin{subfigure}[h]{6.75 cm}
		\includegraphics[width=1.5 cm,height=1.5 cm]{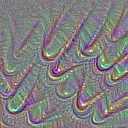}
		\includegraphics[width=1.5 cm,height=1.5 cm]{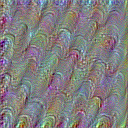}
		\includegraphics[width=1.5 cm,height=1.5 cm]{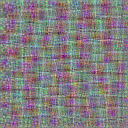}
		\includegraphics[width=1.5 cm,height=1.5 cm]{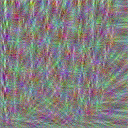}
	\caption{Visualization of layer nine filters}		
	\end{subfigure}
	\end{center}
\caption{Visualization of low level and high level features of VGG}	     
\end{figure}
\newline \indent As we can see from the above figure, the bottom layered layers learn simple features such as different orientations, edges, sharpness etc. But as we go through the different layers more complex features are learned such as particular shape, pattern etc.
\newline \indent The network should learn relevant features from the data. Constraining the network to learn relevant features will reduce the overfitting on the training dataset. Consider the case of classifying the images of cats and dogs. This is a simple task for humans, and in this case we know that the information in the background such as wall, trees etc. are not important and will not be useful while classifying the images into cats and dogs. Every layer in the network learns some features from the data, and if these features are learned predominantly from the background information then the network will not perform well. If this network does perform well on the training data then it will not generalize well as it has not learned what a dog or a cat is. In this case it has learned some other criteria from the background that classifies training data well. We need to ensure that the network learns the relevant features from the data and suppress the background information.
\newline \indent The activation functions in DNN are monotonously increasing function, and the input to a particular node in the DNN is given by,
\begin{equation}
X = W^Tx + b;
\end{equation} 
where $x$ is the input vector at that particular node, and $W^T$ is a weight matrix, and $b$ is the bias at that particular node. Hence the output of the particular node is given,
\begin{equation}
y = f(X)
\end{equation}
where f is a monotonously increasing function. The operation $W^Tx$ is nothing but a dot product between weight vector and the input vector. We need to initialize $W^T$ in such a way that it is maximum when $x$ contains useful information(for e.g. where cat or dog is present) and minimize when $x$ contains noisy/background information(for e.g. where wall or trees etc. are present). The dot product between two vectors is maximum when both the vectors are completely aligned with each other and it is minimum when the vectors are oppositely aligned with each other. Hence to encourage useful features and suppress unuseful features the initialized weight vectors need to be as closely aligned as possible to the useful features. 
\subsection{PCA based initialization}
If a particular feature is important, then it should be present in all the data samples. We make this assumption in all the following cases. If the feature is present in all the data samples, we should be able to get a rough estimate of useful features. Let us use PCA of the data as a rough estimate of data. The algorithm for PCA based initialization is shown below.
\begin{algorithm}[H]
  \begin{algorithmic}[1]
  \FOR{ each image in sample}
  \STATE take the first \textit{$m\times m$} block and then vectorize the block.
  \STATE horizontally stack all the vectors  generated in previous step to form a matrix $U_1$, which will have a size \textit{$N\times mm$}, where N is total number of images
  \STATE Now take the next block in the image i.e. beginning from (1,1) and repeat step 1 and 2. Repeat this process each time moving 1 pixel to the right or down depending on current position,until all blocks have been visited
  \STATE now for $i_th$ block we have a matrix $U_i$ of size $N\times mm$.Now, get the eigenvectors of $U_i^TU_i$ (This matrix has size $mm\times mm$).Now take the eigenvectors from all the blocks and take their average, call this matrix $E$(the columns of E are the average eigenvectors over all the blocks)
  \STATE Repeat above steps for each channel in image and take average of eigenvectors from all channels.
  \ENDFOR
  \end{algorithmic}
  \caption{PCA based initialization}
  \end{algorithm}
\indent We employ the above algorithm on statfarm distracted driver detection challenge, where we need to identify the different actions being performed by the driver such as safely driving, texting, talking on phone or drinking. One such image where the driver is driving safely is shown below.
\begin{figure}[h]
 		\centering
 		\includegraphics[width=0.25\linewidth]{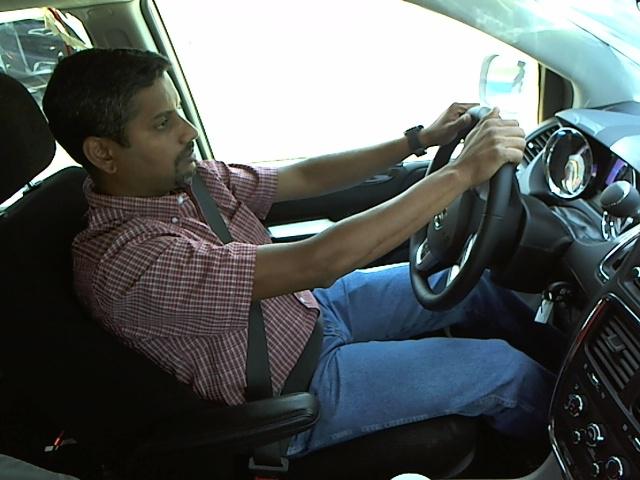}
 		\caption{Original Image \cite{statefarm}}\label{fig:1a}		
\end{figure}
\newline The output of filtering using top 4 PCA filters is shown below
	\begin{figure}[h]
	\centering
		\includegraphics[width=2 cm,height=2cm]{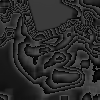}
		\includegraphics[width=2 cm,height=2cm]{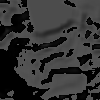}
		\includegraphics[width=2 cm,height=2cm]{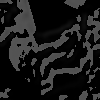}
		\includegraphics[width=2 cm,height=2cm]{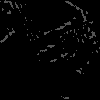}
	\caption{Output of top 4 PCA filters on a general image}\label{fig:1}
\end{figure}
\newline
\indent The PCA does learn the useful information from the data, the background information in the image is suppressed and useful information such as human posture etc. have been highlighted. We initialize the network using the PCA of the data at each layer. And the training profile with this initialization is shown below.
\begin{figure}[h]
\centering
\begin{subfigure}{.5\textwidth}
  \centering
  \includegraphics[width=0.8\textwidth,height=0.2\textheight]{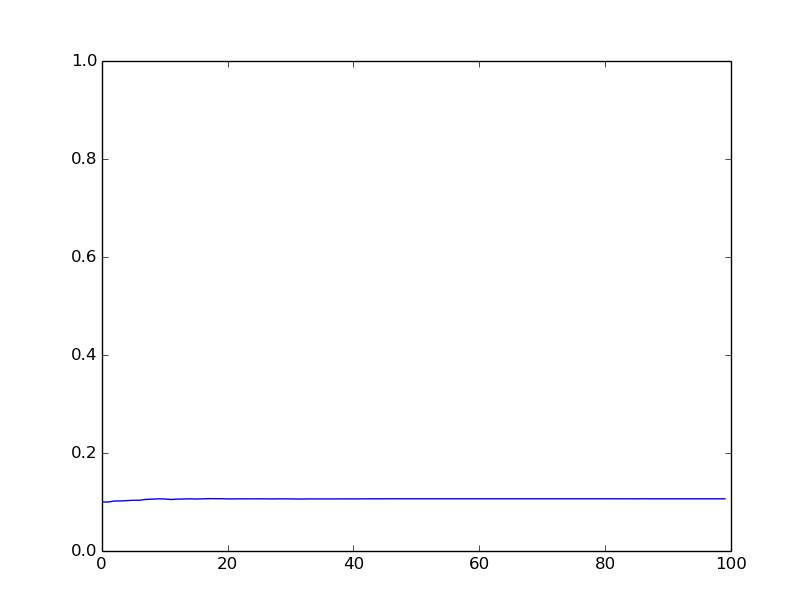}
  \caption{Training Loss}
  \label{fig:sub1}
\end{subfigure}%
\begin{subfigure}{.5\textwidth}
  \centering
  \includegraphics[width=0.8\textwidth,height=0.2\textheight]{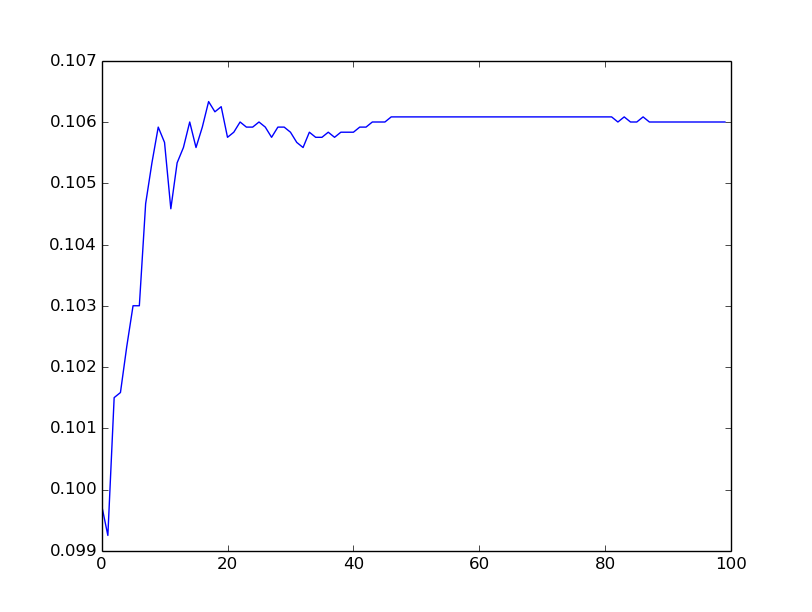}
  \caption{Zoomed in training loss}
  \label{fig:sub2}
\end{subfigure}
\caption{Logloss vs epoch for PCA based initialization}
\label{fig:test}
\end{figure}\par As we can see from the results that, although PCA does learn some useful features from data, but the model initialized using PCA does not converge very well. The reason is PCA does not ensure that all layers learn at the same rate, and The network layers end up not learning anything(The training loss is constant with the no. of epochs). This result emphasizes the importance of Xavier and He initialization.
\subsection{Initialization using data statistics}
\par We saw in the previous section that, despite learning the useful features from the data, PCA based initialization does not converge well. Furthermore, calculating the PCA for such a large dataset takes a lot of time and yields very bad results. Let us try first determine the data statistics and use the samples from the statistics as an estimate of relevant features. At each layer we have
 \begin{equation}
 y_l = W_l^Tx_l + b_l
 \end{equation}
 We want the weights to learn only those features from the data which are responsible for data classification or any other machine learning task. Naturally, these features should be present in all the data samples. Hence, if we choose our network weights to be aligned in the same direction of as that of these particular features, then it will suppress the part of data that yields meaningless features for performing this machine learning task, and will aid the features that are responsible for performing the task. We propose the following algorithm for network weight initialization. Let $n_k$ be number of filters at layer $k$, and $m_k\times m_k$ be the filter size at layer $k$ and $D$ is the training dataset that we have.
 \begin{algorithm}[H]
 \begin{algorithmic}[1]
 \FOR{each affine layer k}
 \STATE Draw samples \boldmath$z_o \in \widetilde{D} \subset D$ and pass through first k-1 layers(for k=1, the output of these layers will be original data)
 \STATE Take n random crops of given filter size from each image
 \STATE Compute the mean$X$ and covariance matrix$C$ from the random crops
 \STATE Initialize weights from \boldmath$W_k \sim N (X, C)$ and biases \boldmath$b_k = 0$ 
 \STATE Whiten the $\textbf{W}_k$ matrix so that it has \textbf{0} mean, and \textbf{\textit{I}}(identity matrix) as it's covariance
 \STATE Scale the elements of $\textbf{W}_k$ such that it has zero mean, and variance as given by He initialization(i.e $\frac{2}{m}$, where m is number of incoming neurons)
 \ENDFOR
 \end{algorithmic}
 \caption{Data dependent weight initialization}
 \end{algorithm}
 \indent In the proposed algorithm we are taking the best of both the worlds, i.e we are making sure that all the layers learn at the same rate as well as making use of data statistics to initialize the network weights so that the weights are roughly aligned to the kind of features that we are trying to capture. This approach is inspired by \cite{krahenbuhl2015data}, but here we are trying to align the weights rather than just scaling the filter weights with the input.In the next section we present the experimental results of this proposed new initialization technique.
 \subsection{Experimental results}
  \par Let us apply the above initialization techniques on few practical and standard datasets and compare our results
  \subsubsection{Bee Image Dataset}
  The given dataset contains two species of bees, namely apis and bombus (as shown in fig. below). The task is to classify each image to their particular species. The features such as color, texture etc. play an important role in classifying these images. 
  \begin{figure}[h]   
  \label{Figure: Bee Species}      
  \begin{center}    
  \includegraphics[scale=0.25]{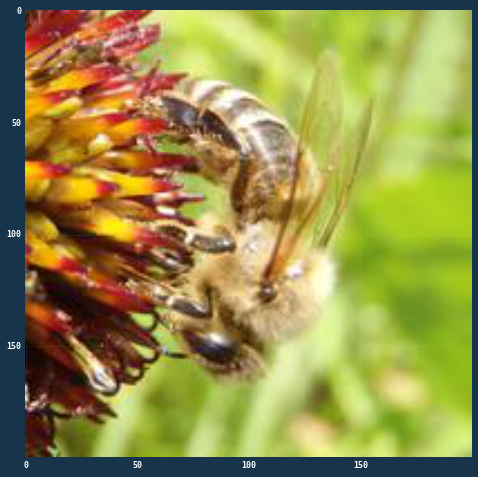} 
  \includegraphics[scale=0.25]{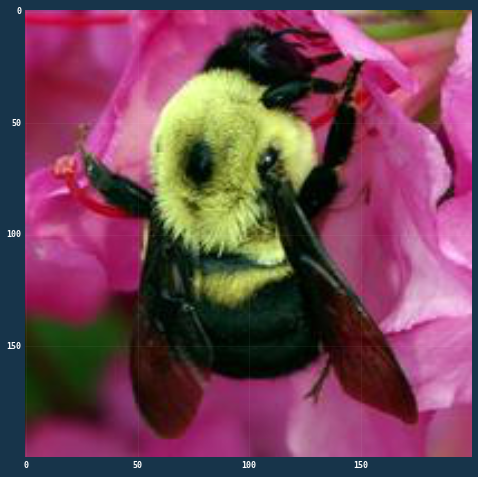} 
  \caption{Bee species: Apis and Bombus\cite{BEE}}   
  \end{center}     
  \end{figure}
  \newpage
  \indent We train the below DCNN architecture for image classification. 
 \begin{figure}[h]   
   \label{Figure: Bee_Arch}      
   \begin{center}    
   \includegraphics[width=8cm,height=3cm]{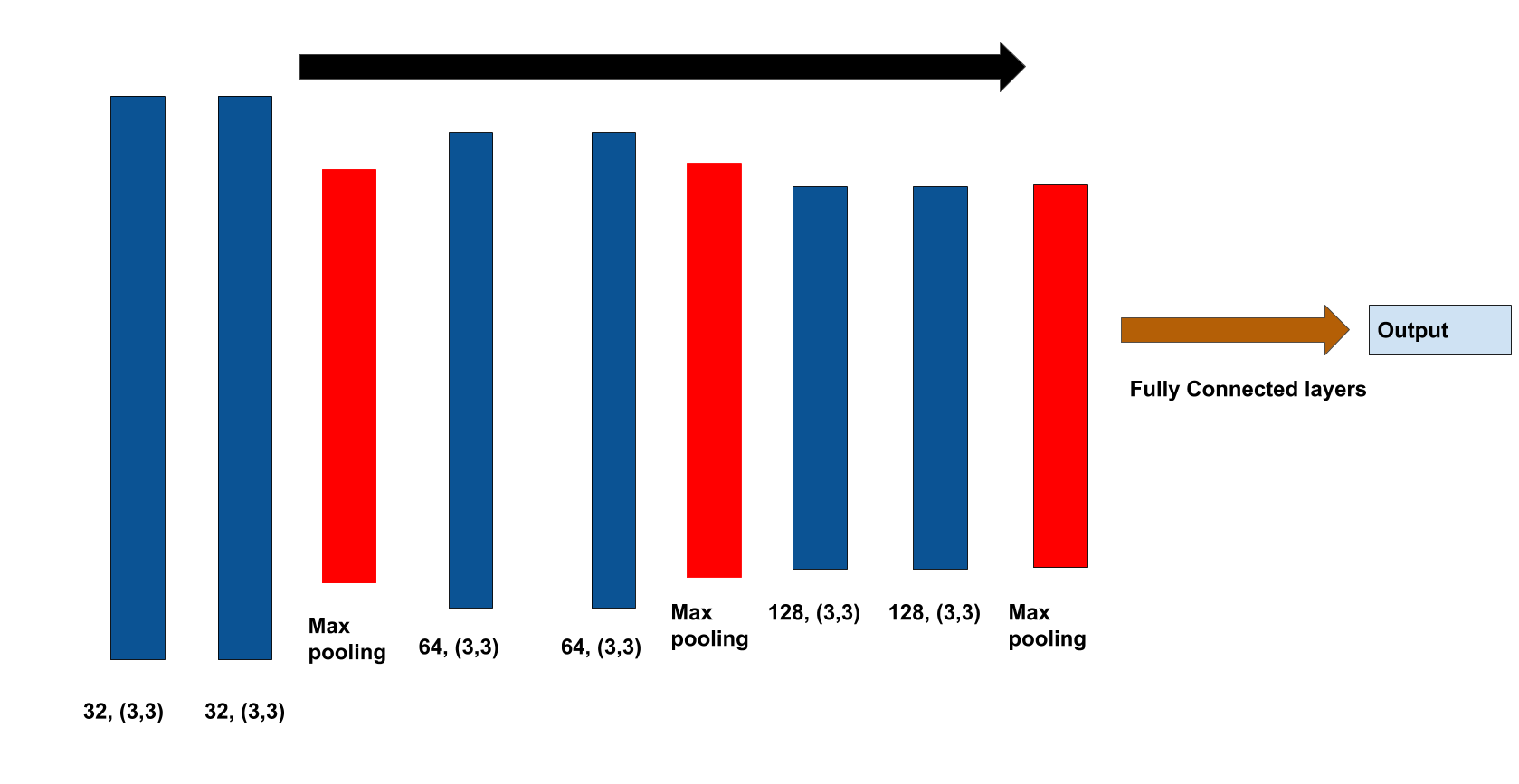} 
   \caption{architecture for bee classification}   
   \end{center}
  \end{figure}
 \newline  
 We initialize the network using both He initialization and the initialization proposed in previous section. The loss profile in the both cases is presented below. 
   \begin{figure}[h]   
    \label{Figure: bee}      
    \begin{center}    
    \includegraphics[width=4cm,height=3cm]{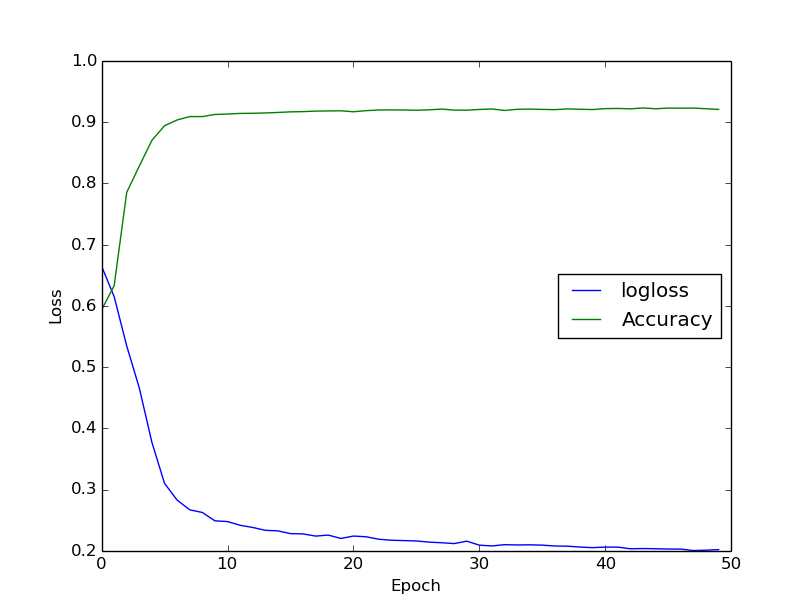}
    \includegraphics[width=4cm,height=3cm]{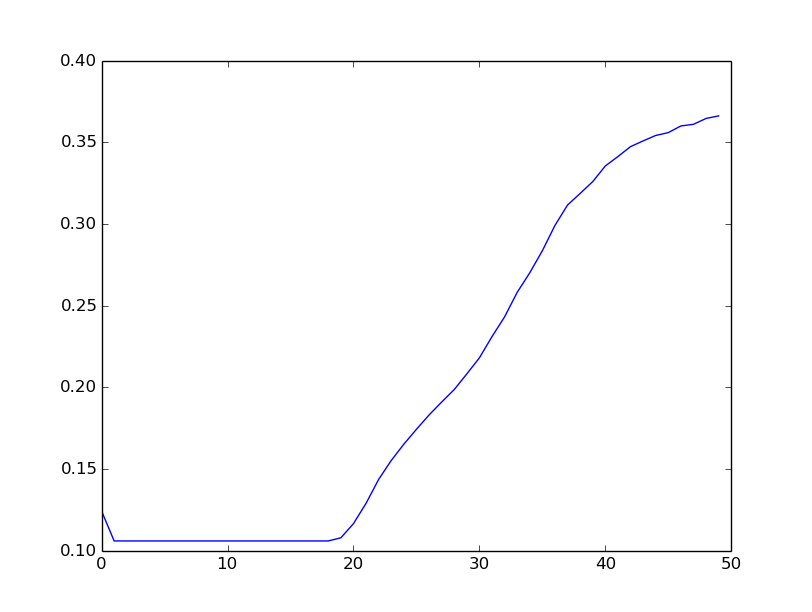}  
    \caption{Accuracy on validation data vs epoch for bee images(left: our method, right: He initialization)}  
    \end{center}     
    \end{figure}
  As we can see from the above figures, after training the network for 50 epoches we achieve an AUC of 0.86 using our initialization whereas the He initialization gets stuck around 0.45 on validation data.
  \subsubsection{Statefarm distracted driver detection}
  This particular dataset consists of the images of drivers performing actions such as safe driving, texting, drinking combing etc. The task is to identify the action being performed in images.
  \begin{figure}[h]
  	\begin{center}
  		\includegraphics[height=1.5cm]{img_34}
  		\includegraphics[height=1.5cm]{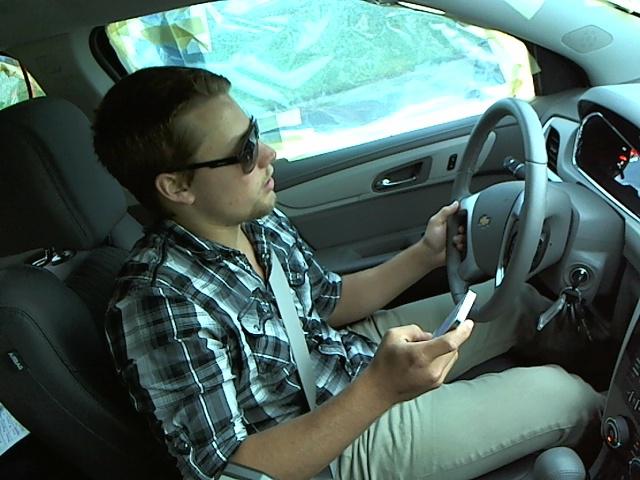}
  		\includegraphics[height=1.5cm]{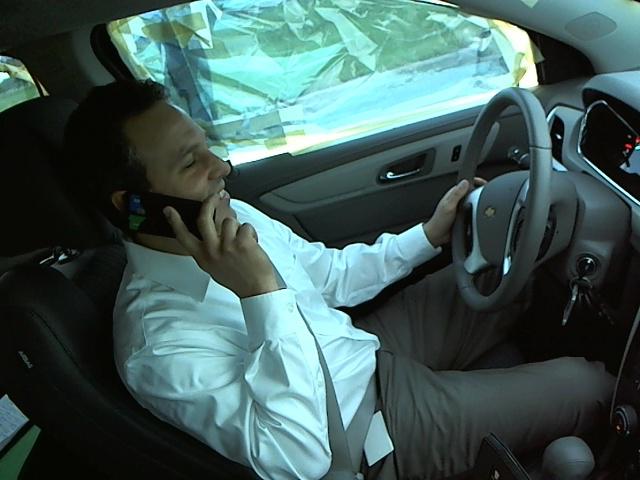}
  		\includegraphics[height=1.5cm]{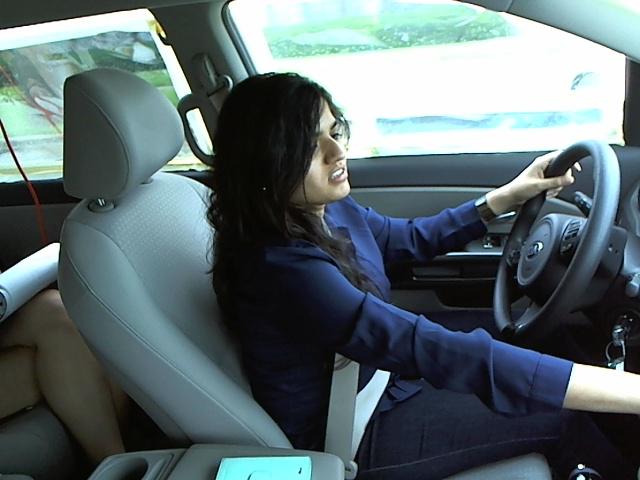}
  	\end{center}
  	\caption{Drivers performing different actions safe driving, texting, talking and operating the radio (from left to right) respectively}\label{fig: 1}
  \end{figure}\newpage
  The architecture used for this particular task is shown below
     \begin{figure}[htpb]   
       \label{Figure:CIFAR_Arch}      
       \begin{center}    
       \includegraphics[width=8 cm,height=3cm]{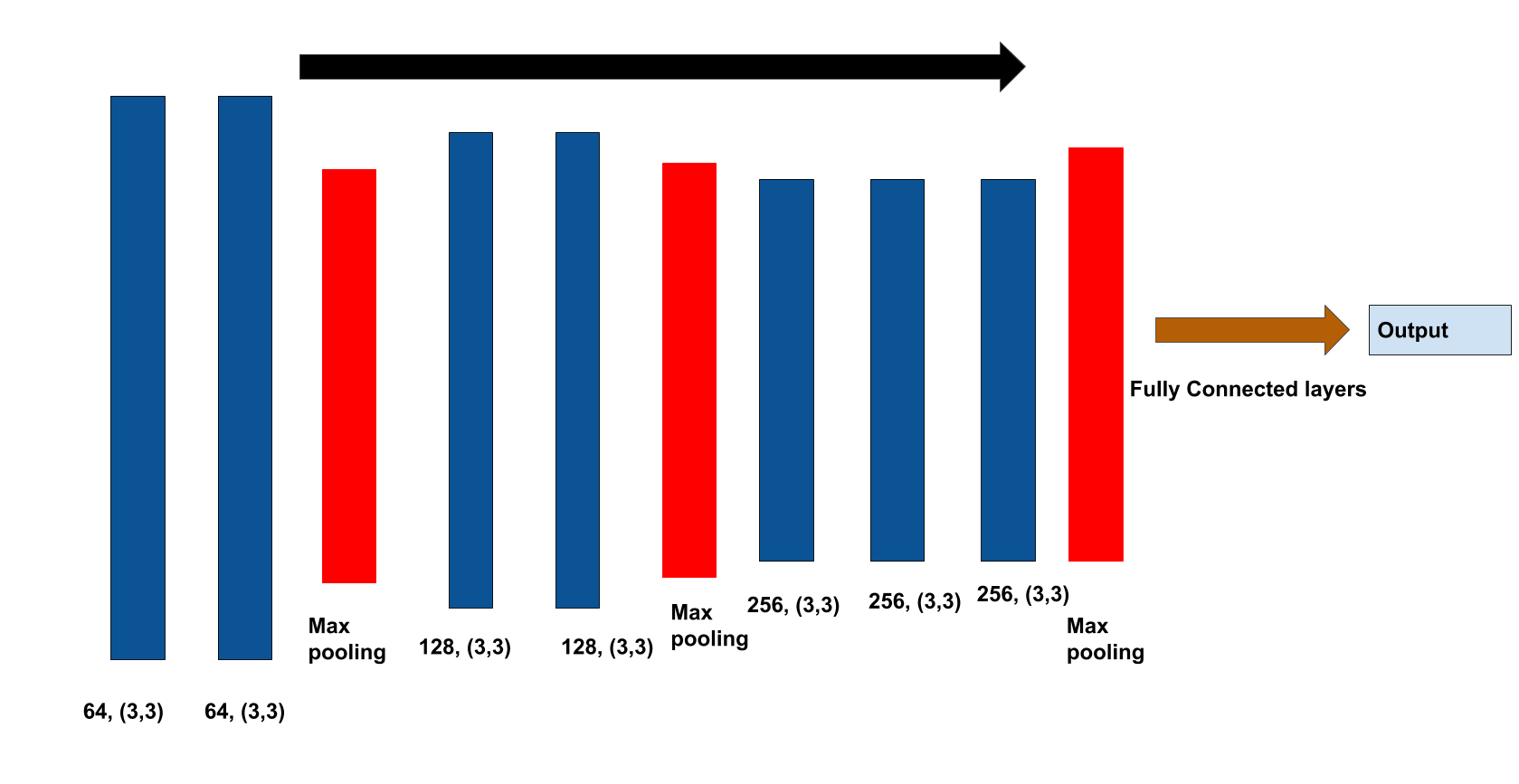} 
       \caption{Statefarm architecture}   
       \end{center}     
       \end{figure} 
  \newline The final performance using both the approaches is shown in fig. 13
   \begin{figure}[htpb]   
    \begin{center}
    \includegraphics[width=4 cm,height=2.6cm]{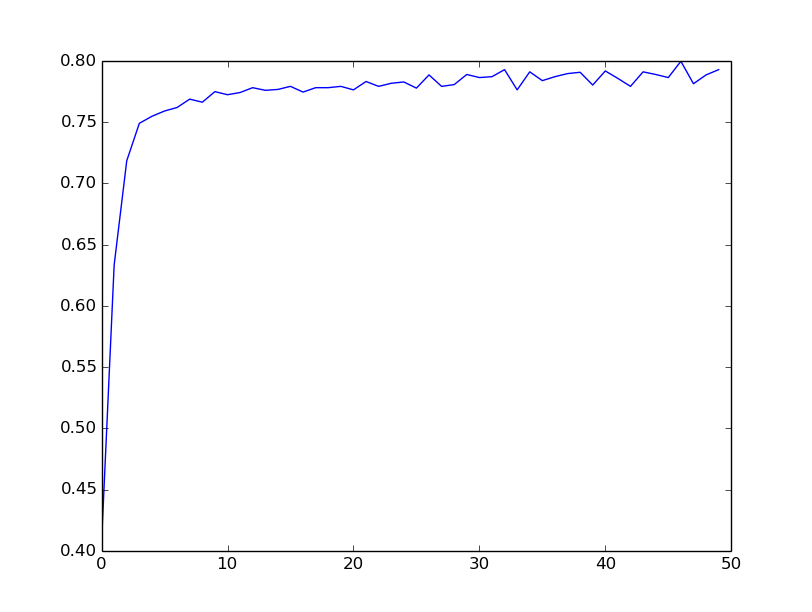}
    \includegraphics[width=4 cm,height=2.6cm]{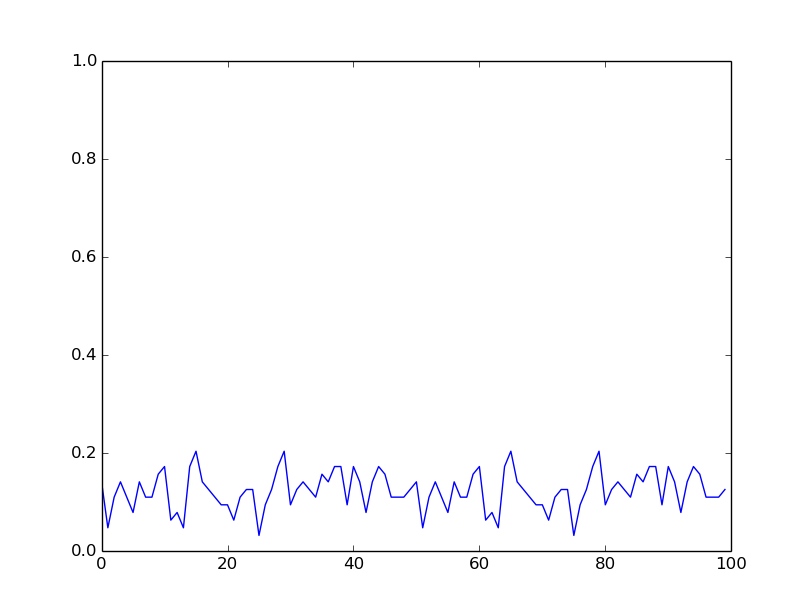}  
    \caption{Accuracy on validation data vs epoch (left: our method, right: He initialization)}  
    \end{center}     
    \end{figure}
  \newpage  
  \subsubsection{Cifar-10}
  This is a balanced dataset which consists $32\times32$ colored images belonging to 10 different classes as shown in fig.14
  \begin{figure}[htpb]   
   \label{Figure: CIFAR}      
   \begin{center}    
   \includegraphics[scale=0.4]{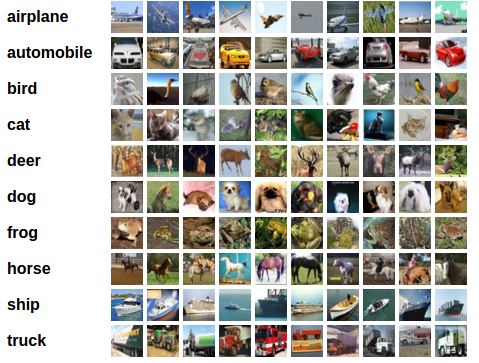} 
   \caption{cifar-10 images\cite{CIFAR}}   
   \end{center}     
   \end{figure}
    \newline Deep neural network architecture 
    \begin{figure}[htpb]   
      \label{Figure: CIFAR_Arch}      
      \begin{center}    
      \includegraphics[width=8 cm,height=3cm]{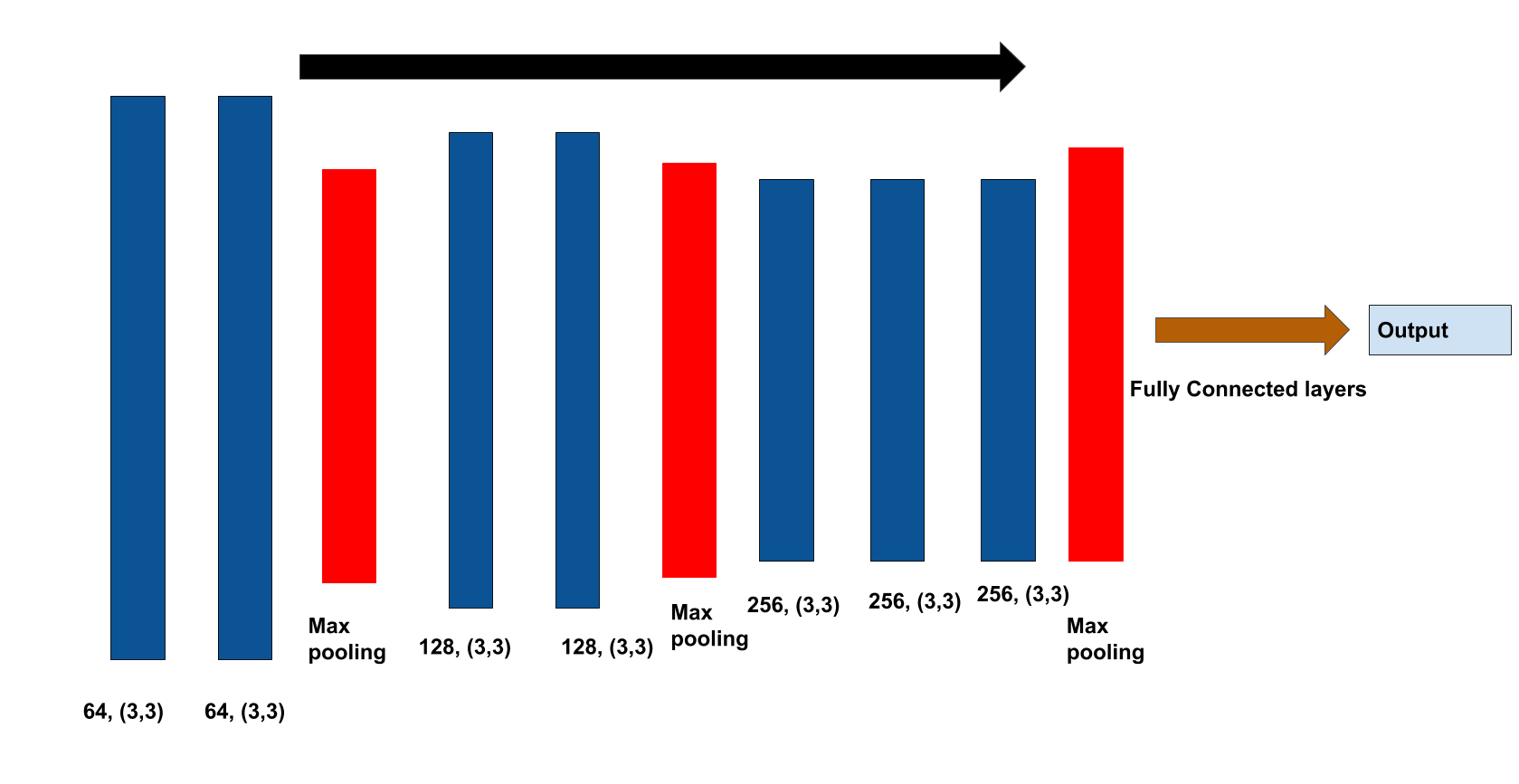} 
      \caption{cifar-10 images\cite{CIFAR}}   
      \end{center}     
      \end{figure}
\newline \indent The final performance using both the initialization techniques
   \begin{figure}[htpb]   
    \label{Figure: statefarm}      
    \begin{center}    
    \includegraphics[width=4 cm,height=3cm]{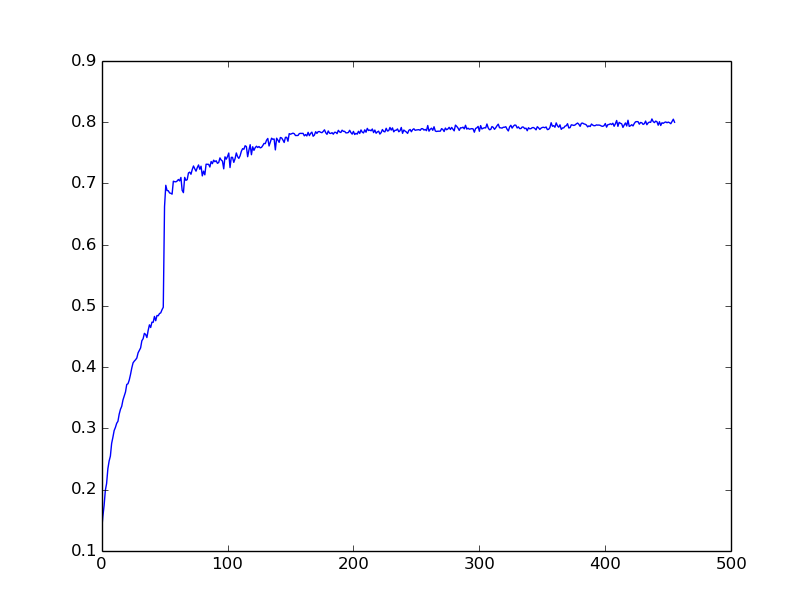}
    \includegraphics[width=4 cm,height=3cm]{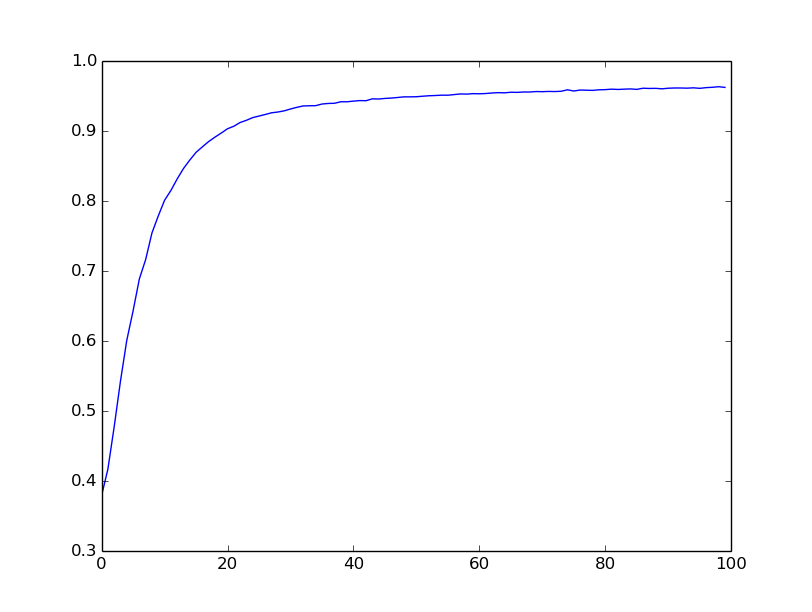}  
    \caption{Accuracy on validation data vs epoch for cifar(left: our method, right: He initialization)}  
    \end{center}     
    \end{figure}
    \newpage        
\subsubsection{MNIST}
   This dataset consists $28\times28$ images of handwritten digits. This is the easiest dataset to work on, even simple SVM with linear kernels also yield accuracy of around 0.9. Few of the images are shown in the following figure.
 \begin{figure}[htpb]   
     \label{Figure: MNIST}      
     \begin{center}    
     \includegraphics[width=6 cm,height=2 cm]{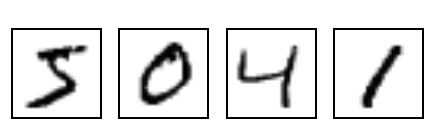} 
     \caption{MNIST images\cite{MNIST}}   
     \end{center}     
 \end{figure}
 \newline \indent We use the following architecture for image classification 
 \begin{figure}[htpb]   
     \label{Figure: MNIST_arch}      
     \begin{center}    
     \includegraphics[width=8 cm,height=3cm]{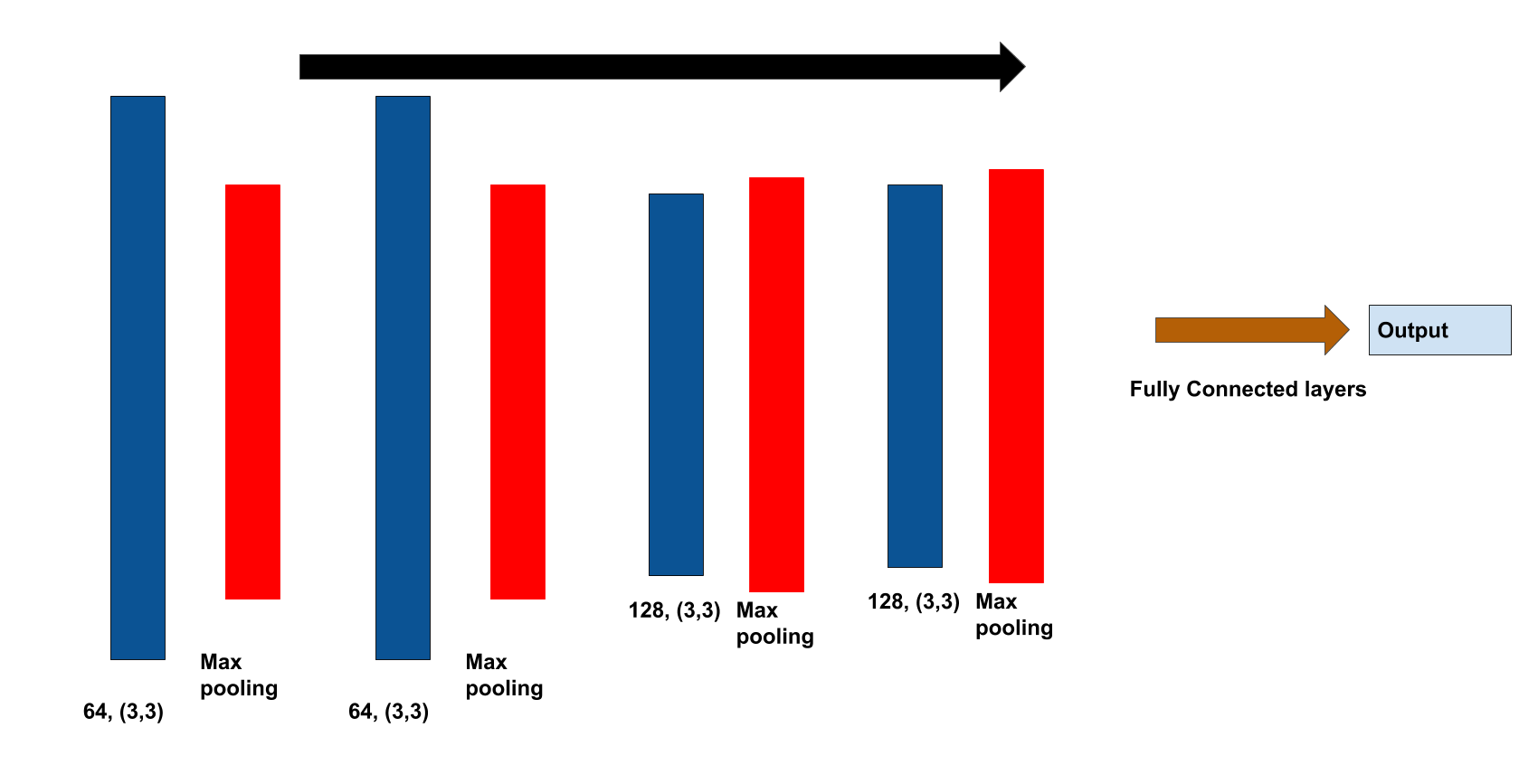} 
     \caption{MNIST architecture}   
     \end{center}     
 \end{figure}
 \newline \indent The final performance is shown below 
 \begin{figure}[htpb]   
      \label{Figure: MNIST performance}      
      \begin{center}    
      \includegraphics[width=4cm,height=3cm]{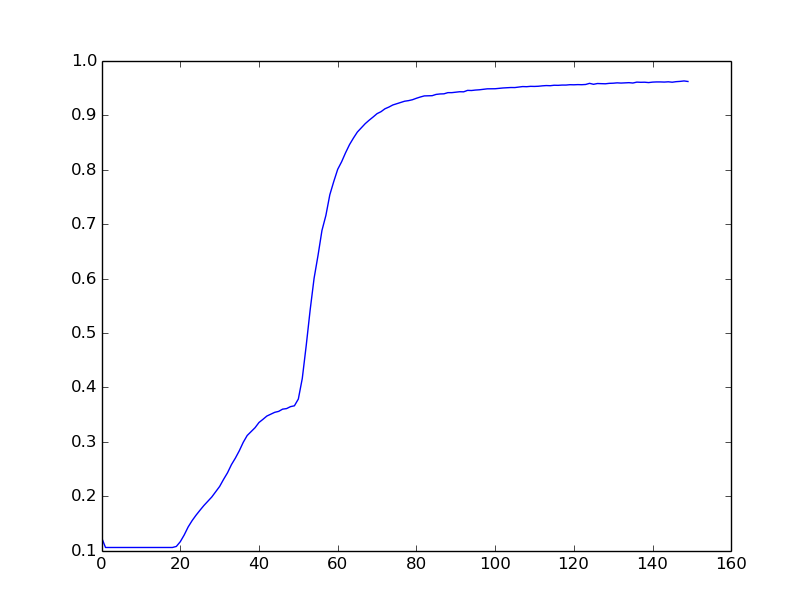} 
      \includegraphics[width=4cm,height=3cm]{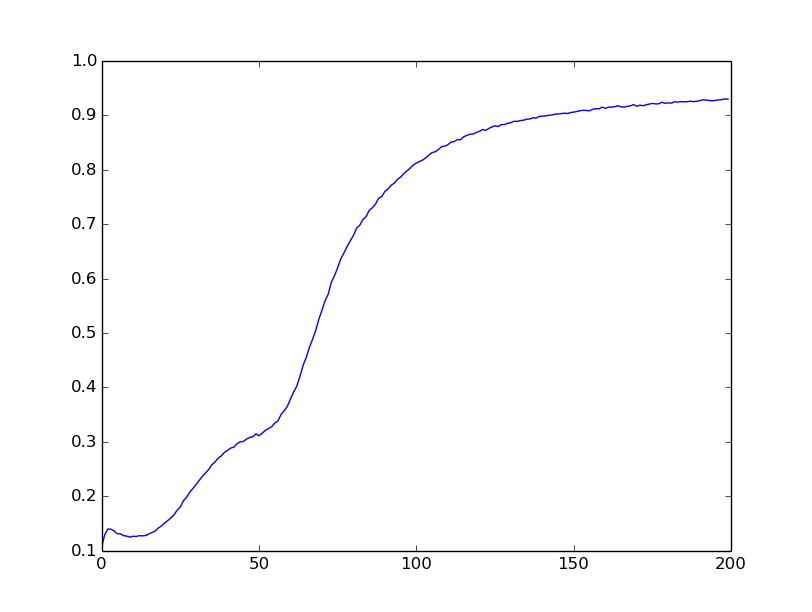} 
      \caption{Accuracy on validation data vs epoch for mnist}  
      \end{center}     
 \end{figure}
 \section{Visualization of learned DNN's}
 \indent Visualization of the deep neural networks gives us insight into how well the network has learned about data. Let us apply this approach on our trained networks and analyse the results.
  \subsection{Bee image network visualization}
  \begin{figure}[h]   
       \label{Figure: bee visualization}      
       \begin{center}    
       \includegraphics[width=4cm,height=3cm]{apis} 
       \includegraphics[width=4cm,height=3cm]{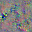} 
       \caption{original bee(left) and visualized bee(right)}  
       \end{center}     
  \end{figure}
  \subsection{Cifar}
  This dataset has very small size ($32\times 32$) hence after visualizing the output, the images look blurred and it is difficult to tell whether the network has learned well or not(see fig.)
  \begin{figure}[h]   
       \label{Figure: cifar visualization}      
       \begin{center}    
       \includegraphics[width=4 cm,height=2.6cm]{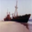} 
       \includegraphics[width=4 cm,height=2.6cm]{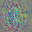} 
       \caption{original ship image(left) and visualized ship(right)}  
       \end{center}     
  \end{figure}
  \newpage
  \subsection{MNIST}
  This is the easiest dataset and our network learns pretty well on this dataset as depicted in the following figure
  \begin{figure}[htpb]   
       \label{Figure: digit visualization}      
       \begin{center}    
       \includegraphics[width=1.5 cm,height=1.5cm]{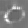} 
       \includegraphics[width=1.5 cm,height=1.5cm]{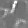} 
  	 \includegraphics[width=1.5 cm,height=1.5cm]{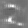} 
       \includegraphics[width=1.5 cm,height=1.5cm]{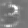}
       \includegraphics[width=1.5 cm,height=1.5cm]{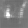}  
       \caption{Visualized handwritten digits}  
       \end{center}     
  \end{figure}\par
  \begin{figure}[htpb]   
       \label{Figure: digit visualization}      
       \begin{center}    
       \includegraphics[width=1.5 cm,height=1.5cm]{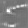} 
       \includegraphics[width=1.5 cm,height=1.5cm]{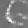} 
  	 \includegraphics[width=1.5 cm,height=1.5cm]{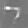} 
       \includegraphics[width=1.5 cm,height=1.5cm]{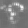}
       \includegraphics[width=1.5 cm,height=1.5cm]{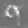}  
       \caption{Visualized handwritten digits}  
       \end{center}     
  \end{figure}
  \newpage
  \subsection{Statefarm distracted driver detection}
  The output visualization on final network
  \begin{figure}[h]
  	\begin{center}
  		\includegraphics[width=2 cm,height=2cm]{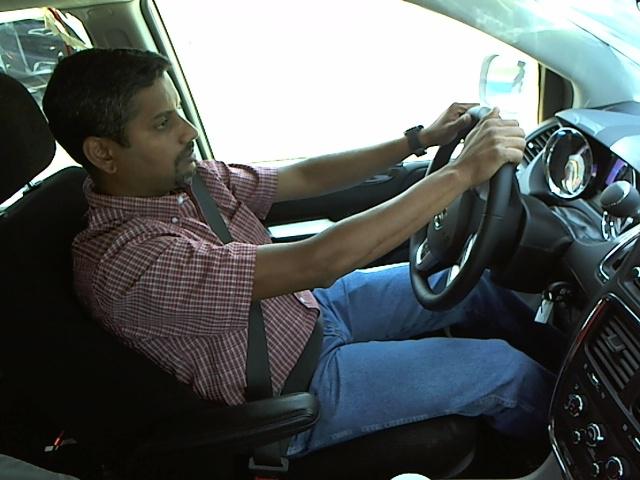}
  		\includegraphics[width=2 cm,height=2cm]{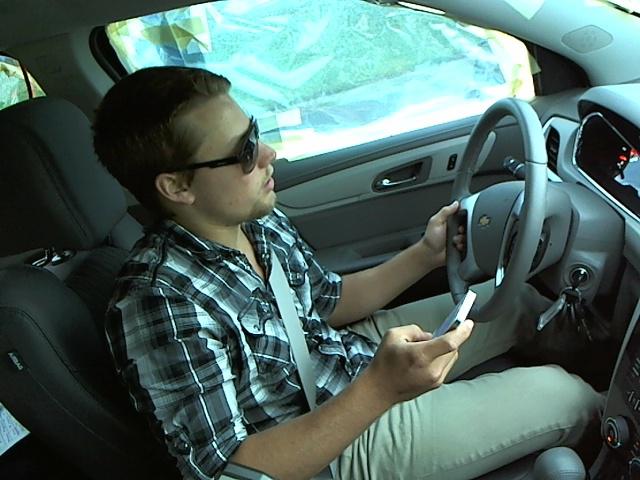}
  		\includegraphics[width=2 cm,height=2cm]{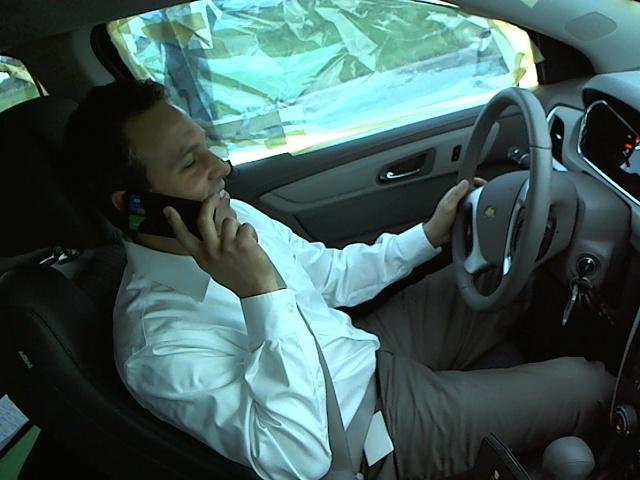}
  		\includegraphics[width=2 cm,height=2cm]{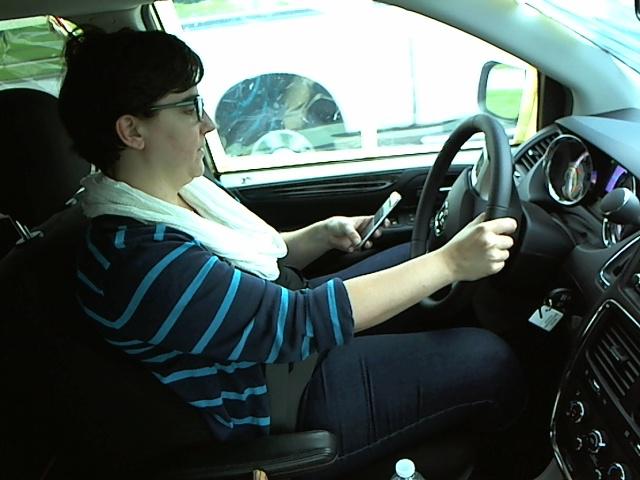}
  		\\
  		\includegraphics[width=2 cm,height=2cm]{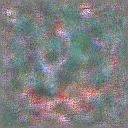}
  		\includegraphics[width=2 cm,height=2cm]{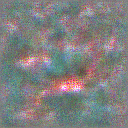}
  		\includegraphics[width=2 cm,height=2cm]{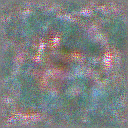}
  		\includegraphics[width=2 cm,height=2cm]{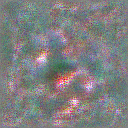}
  	\end{center}
  	\caption{Original images(top) and the visualized images(bottom)-I}\label{fig:1}
  \end{figure}
  
  \begin{figure}[htpb]
  \begin{center}    
  		\includegraphics[width=2 cm,height=2cm]{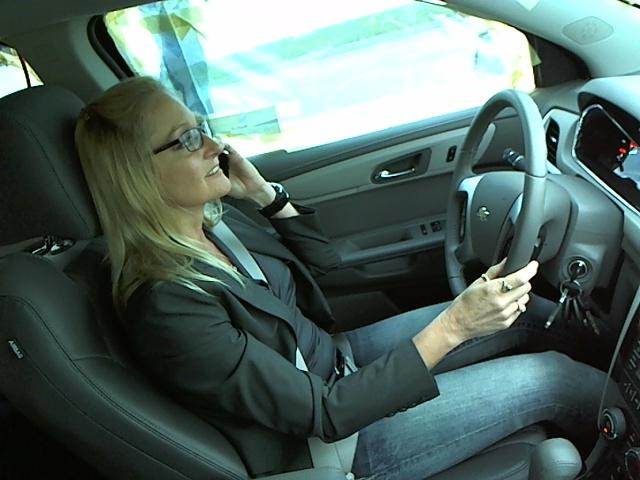}
  		\includegraphics[width=2 cm,height=2cm]{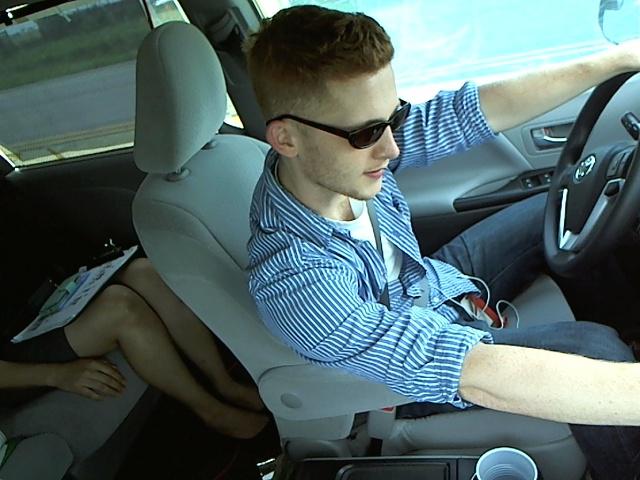}
  		\includegraphics[width=2 cm,height=2cm]{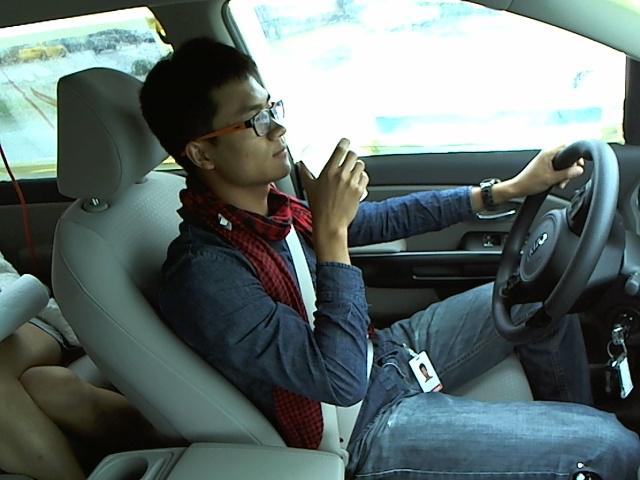}
  		\includegraphics[width=2 cm,height=2cm]{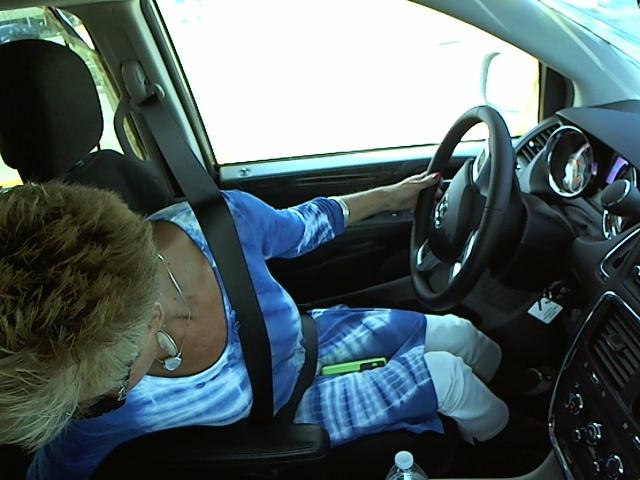}
  		\\
  		\includegraphics[width=2 cm,height=2cm]{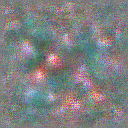}
  		\includegraphics[width=2 cm,height=2cm]{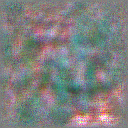}
  		\includegraphics[width=2 cm,height=2cm]{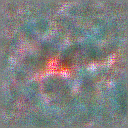}
  		\includegraphics[width=2 cm,height=2cm]{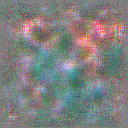}
  	\caption{Original images(top) and the visualized images(bottom)-II}\label{fig:1}
  	\end{center}     
  \end{figure}
  \newpage
  \begin{figure}[h]
   	\centering
  	\begin{subfigure}[h]{5 cm}
  		\includegraphics[width=2 cm,height=2cm]{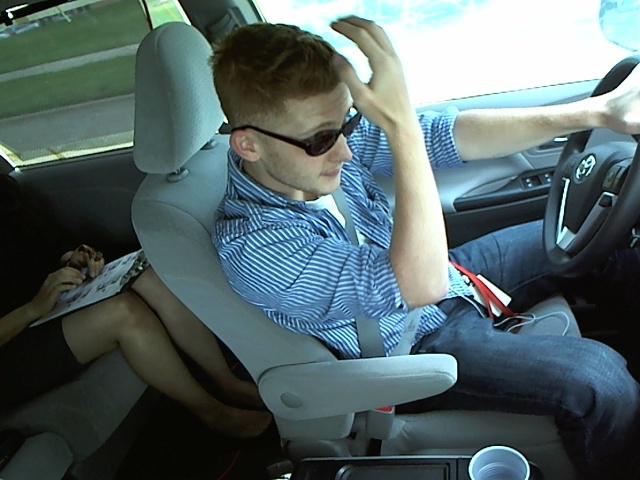}
  		\includegraphics[width=2 cm,height=2cm]{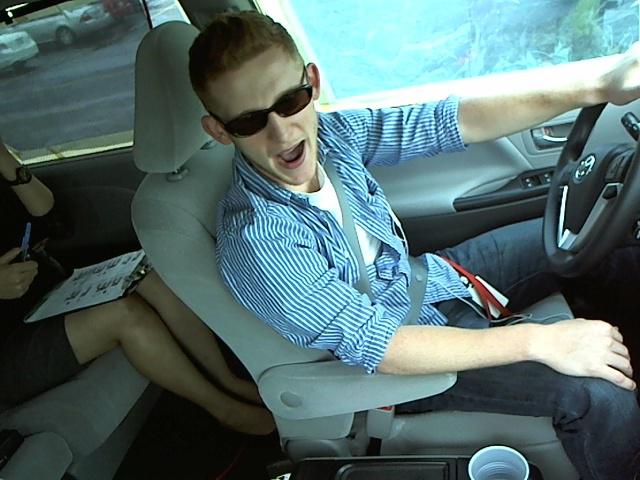}
  		\\
  		\\
  		\includegraphics[width=2 cm,height=2cm]{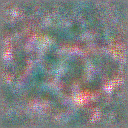}
  		\includegraphics[width=2 cm,height=2cm]{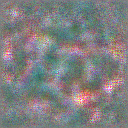}
  	\end{subfigure}
  	\caption{Original images(top) and the visualized images(bottom)-II}\label{fig:1}
  \end{figure}
  As can be seen from the above table that the heatmap of visualized images highlights the appropriate places in the image, this is a good indication that our network has learned well about the data.
 \section{Results and discussions}
   \indent One can see from the previous results that standard weight initialization techniques such as He and glorot perform very poorly on the practical datasets such as bee images and statefarm distracted driver detection. Whereas our initialization technique which uses data statistics performs quite well on practical datasets. These results further emphasize the need of using data statistics for network weight initialization. Although it is important to keep the learning rate across the layers to be constant, but ensuring only this condition does not give good results. The approach suggested by us takes the better of both the worlds and initialize network weights closer to useful features from the data. And our results confirm that our approach works well on standard datasets and also achieve the good results on practical datasets.
   \section{Acknowledgemnt}
   \indent I would like to thank the Department of Electrical Engineering, IIT Bombay for granting me access to the SPANN lab and it's HPC server, most of the work would have been incomplete without it. I would also like to thank the keras\cite{keras} community. Almost each and every experiment has been programmed in python using keras deep learning library. Keras has been a big help for my work and none of this would have been possible without keras. I would also like to thank the drivendata comunity\cite{BEE} and statefarm community\cite{Kaggle} for letting us use the data for research purpose.


\bibliographystyle{elsarticle-num}
\bibliography{report}

\begin{thebibliography}{10}
\expandafter\ifx\csname url\endcsname\relax
  \def\url#1{\texttt{#1}}\fi
\expandafter\ifx\csname urlprefix\endcsname\relax\def\urlprefix{URL }\fi
\expandafter\ifx\csname href\endcsname\relax
  \def\href#1#2{#2} \def\path#1{#1}\fi

\bibitem{VGG}
K.~Simonyan, A.~Zisserman, Very deep convolutional networks for large-scale
  image recognition, arXiv preprint arXiv:1409.1556.

\bibitem{speech}
G.~Hinton, L.~Deng, D.~Yu, G.~E. Dahl, A.-r. Mohamed, N.~Jaitly, A.~Senior,
  V.~Vanhoucke, P.~Nguyen, T.~N. Sainath, et~al., Deep neural networks for
  acoustic modeling in speech recognition: The shared views of four research
  groups, IEEE Signal Processing Magazine 29~(6) (2012) 82--97.

\bibitem{sentiment}
X.~Glorot, A.~Bordes, Y.~Bengio, Domain adaptation for large-scale sentiment
  classification: A deep learning approach, in: Proceedings of the 28th
  international conference on machine learning (ICML-11), 2011, pp. 513--520.

\bibitem{segemnent}
L.-C. Chen, G.~Papandreou, I.~Kokkinos, K.~Murphy, A.~L. Yuille, Semantic image
  segmentation with deep convolutional nets and fully connected crfs, arXiv
  preprint arXiv:1412.7062.

\bibitem{YOLO}
J.~Redmon, S.~Divvala, R.~Girshick, A.~Farhadi, You only look once: Unified,
  real-time object detection, in: Proceedings of the IEEE Conference on
  Computer Vision and Pattern Recognition, 2016, pp. 779--788.

\bibitem{sepp}
S.~Hochreiter, Untersuchungen zu dynamischen neuronalen netzen, Ph.D. thesis,
  diploma thesis, institut f{\"u}r informatik, lehrstuhl prof. brauer,
  technische universit{\"a}t m{\"u}nchen (1991).

\bibitem{pretraining}
D.~Erhan, Y.~Bengio, A.~Courville, P.-A. Manzagol, P.~Vincent, S.~Bengio, Why
  does unsupervised pre-training help deep learning?, Journal of Machine
  Learning Research 11~(Feb) (2010) 625--660.

\bibitem{glorot}
X.~Glorot, Y.~Bengio, Understanding the difficulty of training deep feedforward
  neural networks., in: Aistats, Vol.~9, 2010, pp. 249--256.

\bibitem{He}
K.~He, X.~Zhang, S.~Ren, J.~Sun, Delving deep into rectifiers: Surpassing
  human-level performance on imagenet classification, in: Proceedings of the
  IEEE international conference on computer vision, 2015, pp. 1026--1034.

\bibitem{statefarm}
Statefarm distracted driver detection,
  \url{https://www.kaggle.com/c/state-farm-distracted-driver-detection/leaderboard},
  accessed: 2017-06-14.

\bibitem{krahenbuhl2015data}
P.~Kr{\"a}henb{\"u}hl, C.~Doersch, J.~Donahue, T.~Darrell, Data-dependent
  initializations of convolutional neural networks, arXiv preprint
  arXiv:1511.06856.

\bibitem{BEE}
Naive bees classifier,
  \url{https://www.drivendata.org/competitions/8/naive-bees-classifier/},
  accessed: 2017-06-14.

\bibitem{CIFAR}
Cifar-10 dataset,
  \url{http://prog3.com/sbdm/blog/cyh_24/article/details/50593400}, accessed:
  2017-06-14.

\bibitem{MNIST}
Cifar-10 dataset, \url{http://yann.lecun.com/exdb/mnist}, accessed: 2017-06-14.

\bibitem{keras}
F.~Chollet, et~al., Keras (2015).

\bibitem{Kaggle}
Diabetic retinopathy winner interview,
  \url{http://blog.kaggle.com/2015/09/09/diabetic-retinopathy-winners-interview-1st-place-ben-graham/},
  accessed: 2017-06-14.

\end{thebibliography}

\end{document}